\documentclass[10pt,twocolumn,letterpaper]{article}

\usepackage{iccv}              
\usepackage{graphicx}
\usepackage{amsmath}
\usepackage{amssymb}
\usepackage{booktabs}
\usepackage{adjustbox}
\usepackage{tabularx}
\usepackage{float}
\usepackage{lipsum}
\usepackage{mathtools}
\usepackage{cuted}
\usepackage{algorithm}
\usepackage{algorithmic}
\usepackage[numbers]{natbib}

\definecolor{iccvblue}{rgb}{0.21,0.49,0.74}
\usepackage[pagebackref,breaklinks,colorlinks,allcolors=iccvblue]{hyperref}

\def\paperID{} 
\def\confName{ICCV}
\def\confYear{2025}

\title{Leveraging Diffusion Models for Stylization using Multiple Style Images}

\author{Dan Ruta~~~~Abdelaziz Djelouah~~~~Raphael Ortiz~~~~Christopher Schroers\\  ~ \vspace{-0.3cm}\\
DisneyResearch\textbar Studios\\ ~ \vspace{-0.3cm}\\
{\tt\small dan.ruta@disney.com~~abdelaziz.djelouah@disney.com}
}

\begin{document}

\twocolumn[{%
	\renewcommand\twocolumn[1][]{#1}%
	\maketitle
	\vspace{-0.5cm}
	\hspace*{-0.32cm}
	\includegraphics[width=1.01\linewidth]{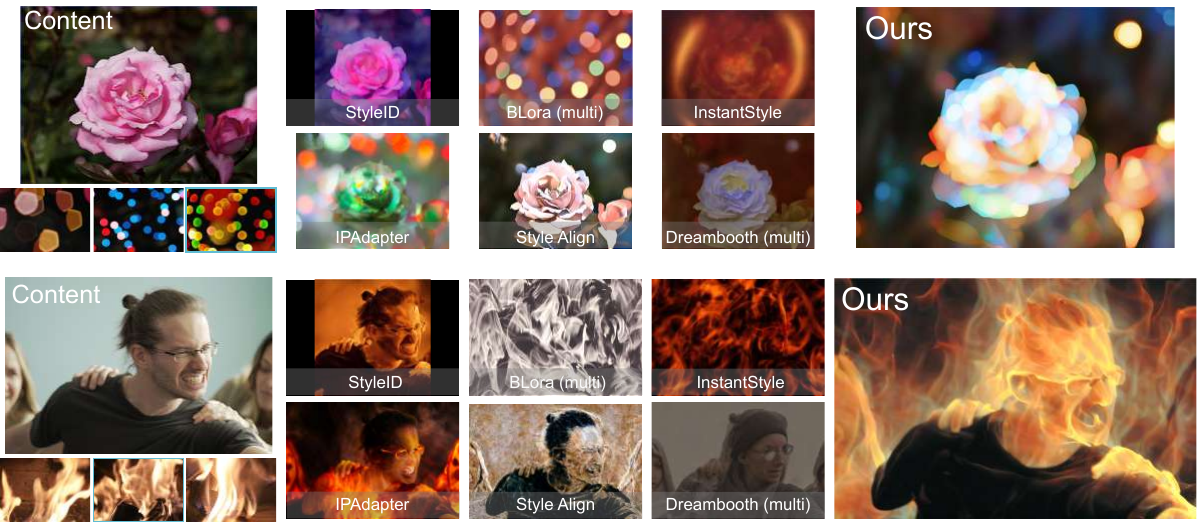}	
	\captionof{figure}{
	We propose a style transfer method that uses multiple style images and achieves state-of-the-art results.
	In each case, our result preserves the content
	while closely matching the style. 
	Existing methods struggle to achieve both content preservation
	and high quality stylization, even the ones using 
	multiple style images such as Dreambooth~\cite{dreambooth} and BLora~\cite{blora}.
	Besides our result, we indicate with \emph{(multi)} the methods using multiple images.
	We highlight in cyan the style image provided for single image based stylization methods.
	}
	\label{fig:teaser}
	\vspace{0.40cm}
}]

\begin{abstract}
Recent advances in latent diffusion models have enabled exciting progress in image style transfer. 
However, several key issues remain. For example, existing methods still struggle to accurately 
match styles. They are often limited in the number of style images that can be used. 
Furthermore, they tend to entangle content and style in undesired ways. 
To address this, we propose leveraging multiple style images 
which helps better represent style features and 
prevent content leaking from the style images.
We design a method that leverages both image prompt adapters 
and statistical alignment of the features during the denoising process. 
With this, our approach is designed such that it can intervene 
both at the cross-attention \emph{and} the self-attention layers of the denoising UNet.  
For the statistical alignment, we employ clustering to distill 
a small representative set of attention features from the large number of attention values extracted from the style samples. As demonstrated in our experimental section, the resulting method achieves state-of-the-art results for stylization.
\end{abstract}
\vspace{-0.3cm}    
\section{Introduction}
\label{sec:intro}

Artists are in constant exploration 
to create new artistic renderings, 
that can offer fresh and different looks.
In this context, image style transfer aims at 
simplifying style exploration with the objective 
of allowing faster iteration during this artistic 
research phase.

Recently, diffusion based image stylization
methods~\cite{stylealigned,style_inject,ipadapter}
have shown impressive results.
For example, image prompt adaptation methods~\cite{ipadapter}
use style information derived from CLIP-image embeddings.
To limit content and style entanglement,
it is possible to use statistical alignment~\cite{stylealigned,style_inject} 
of the content and style image during the denoising process.
In the context of multiple style images, 
personalization methods such as
Dreambooth~\cite{dreambooth}, Lora~\cite{lora2021} or 
CustomDiffusion~\cite{customdiffusion}
can leverage the available style samples for fine-tuning 
(albeit with different strategies).
In all these mentioned works, we observe two issues clearly visible in Figure~\ref{fig:teaser}:
entanglement of content and style, and lower quality style transfer.

Building on insights from recent works, we design a method that
achieves both better content preservation
and higher quality stylization, 
using several style images to address the aforementioned problems.
Our approach can be summarized in the following $3$ steps:
\emph{First}, we fine-tune an image prompt adapter
model on the style images. To help disentangle style from content, we compute
an average token vector from the style images that will be used as prompt
for the diffusion model. 
\emph{Second}, we distill style features from multiple 
images through a clustering approach.
\emph{Finally}, we adopt a two-stage strategy to 
achieve high quality results with some control
on the structural level at which the stylization happens.

Our contributions are the following:
\begin{itemize}
	\item A method that combines model adapters and statistics alignment;
	\item A solution to scale up to multiple style images;
	\item State-of-the-art stylization results as demonstrated in our thorough evaluation, including a user study.
\end{itemize}

\section{Related work}
\label{sec:related}

Image style transfer remains a fundamental challenge in computer vision, aiming to modify the appearance of a content image based on a given reference. Our discussion here mostly focuses on recent works, in particular the ones using latent diffusion models. For a more detailed overview, we refer to the review on style transfer from Jing~\emph{et al.}~\cite{review2019}.

\paragraph{Optimization based stylization.}
Early works such as the seminal Gatys style transfer algorithm~\cite{Gatys_2016_CVPR} rely on inference-time optimization to achieve style transfer. As this is generally an impractical time and resource consuming process, the field has focused research efforts on fast zero-shot approaches.
Still recent work~\cite{nnst} circled back to an iterative approach and achieved some of the best style transfer results.

\paragraph{Using Multiple Images for style transfer.}
Recent works have extended NST to incorporate multiple style references, facilitating better style interpolation and mixing. Some approaches~\cite{huang2019style_mixer, liu2021multiple, wang2023multi} focus on learning a robust style representation capable of blending multiple sources seamlessly. Others employ generative adversarial networks (GANs) \cite{karras2019style}, which, when trained on small datasets, allow for rapid style adaptation through fine-tuning \cite{karras2020training, ojha2021few}. Despite their success in domain-specific applications such as facial stylization, these methods often struggle with generalization beyond constrained settings. In contrast, diffusion-based models have emerged as a powerful alternative for achieving high-quality, diverse style transfer.

\paragraph{Statistical alignment and moment matching.}
Style representation can also be captured through statistical properties of images. Early work introduced Adaptive Instance Normalization (AdaIN) \cite{adain} and Whitening and Coloring Transform (WCT) \cite{wct}, which align the mean and variance of content and style features to achieve stylization. More recent techniques extend this paradigm by incorporating higher-order moments (e.g., skewness and kurtosis) to enhance fidelity in style transfer~\cite{moment_matching, aladinnst}. These methods provide explicit control over style attributes, complementing the implicit representations learned by deep networks.

\paragraph{Diffusion Based.}
Latent diffusion models (LDMs) \cite{ldm} have recently revolutionized style transfer, offering three primary strategies: customization, adaptation modules, and feature alignment. Customization techniques, such as DreamBooth \cite{dreambooth} and Custom Diffusion~\cite{customdiffusion}, fine-tune all model parameters to encode new styles, achieving high-fidelity results at the cost of computational efficiency. Low-Rank Adaptation (LoRA)~\cite{lora2021} mitigates some of these inefficiencies by introducing lightweight fine-tuning mechanisms. Alternatively, adaptation modules condition pre-trained diffusion models on external style information. IP-Adapter~\cite{ipadapter}, for instance, transforms CLIP~\cite{clip} embeddings of style images into inputs compatible with diffusion models, enabling zero-shot style transfer. However, these methods often suffer from content leakage, as they struggle to disentangle content and style effectively.

Feature alignment methods offer a training-free approach to style transfer by manipulating the self-attention layers of diffusion models. DIFF-NST~\cite{diffnst} and Style Injection~\cite{style_inject} replace attention values from the generated image with those from a style reference, effectively transferring texture and color characteristics. StyleAligned~\cite{stylealigned} refines this process by incorporating statistical AdaIN operations within attention layers, ensuring robust style adaptation.

Recent advancements aim to improve content-style disentanglement within diffusion models. BLoRA~\cite{blora} explicitly separates content and style representations using SDXL~\cite{sdxl} and LoRA-based fine-tuning. InstantStyle~\cite{instantstyle} takes a similar approach but introduces feature decoupling and targeted injection into specific layers, reducing content leakage and preserving style fidelity.

\begin{figure*}[ht]
	\centering
	\includegraphics[width=2.1\columnwidth]{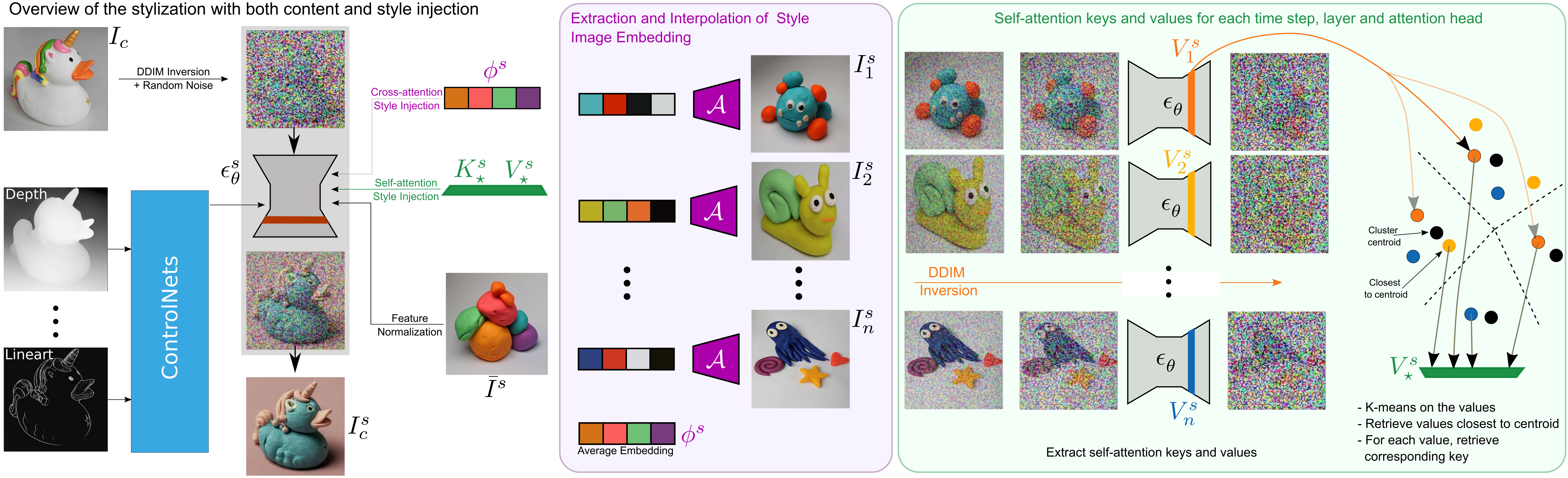}
	\caption{Overview of the diffusion based stylization method. \textbf{(Left)} Given a content image $I_c$ we extract line 
		art and depth maps to guide the content during denoising with ControlNets. 
		Style is enforced at the cross and self attention layers. 
		We use the average style embedding $\phi^s$, representative attention maps keys and values ($K^s_\star, V^s_\star$)
		and an average style image $\bar{I}^s$. 
		\textbf{(Middle)} From the set of input style images $\{I^{s}_1, \ldots, I^{s}_n\}$,
		we train the image prompt adapter $\mathcal{A}$. The average embedding $\phi^s$
		is obtained by interpolating the style image embeddings.
		\textbf{(Right)} We use DDIM to invert each style image and extract keys and values for each layer
		and time step. We use k-means clustering to reduce the keys and values to a manageable size,
		keeping the ones closest to the cluster centroid. As keys and values are paired, we perform the clustering on the values only, 
		and retrieve their matching keys. }
	\label{fig:method}
	\vspace{-0.3cm}
\end{figure*}

\section{Multi-Image Style Transfer}

Given a set of style images $\mathcal{S} = \{I^{s}_1, \ldots, I^{s}_n\}$,
our objective is to transform a content image $I_c$ into its stylized
version $I^s_c$, that matches as well as possible the style characteristics
while maintaining the original content.
Our stylization method leverages pre-trained diffusion models.
Specifically we use the latent diffusion model proposed
by Rombach~\emph{et al.}~\cite{ldm}, where an image $I$
is encoded using a variational auto-encoder
into its latent representation $x$.
Diffusion and denoising are done on this latent representation,
where a denoising model is trained to estimate the noise on $x_t$
at each given step $t$, with $\tau_\theta$ transforming the text prompt $y$ and $\theta$ denoting the parameters of the model.
\begin{equation}
	\epsilon_t = \epsilon_\theta(x_t, \tau_\theta(y), t)
\end{equation}

We express our stylization problem in the same framework,
by making the adjustments needed to the denoising model,
adding both content/style images: $I_c$, and $\mathcal{S}$
\begin{equation}
	\epsilon_t = \epsilon^s_\theta(x_t, \tau_\theta(y), t, I_c, \mathcal{S}).
\end{equation}

We illustrate the adjustments we make to the denoising model $\epsilon^s_\theta$
on the left side of Fig.~\ref{fig:method}. We use ControlNets 
to provide a content related signal to the denoiser using
depth and linart features extracted from $I_c$.
Regarding style, we provide guidance at 2 different levels: 
at the \emph{cross-attention} level, with an average style prompt embedding $\phi^S$;
at the \emph{self-attention} level with style representative
keys $K^\star_s$  and values $V^\star_s$ used in the self-attention mechanism,
that we normalize using the average style image $\bar{I}^s$.
In the next subsections we present
these different aspects of the method in more detail,
 with corresponding visualizations in Fig~\ref{fig:method}.

\subsection{Prompt Adaptation with Multiple Style Images}
With multiple style images, it becomes possible to 
mitigate the issues observed with image adapters. 
We propose $2$ adjustments: 
First, the fine-tuning of the image prompt adapter model on the style images.
Second, the interpolation of the style image embeddings.

\paragraph{Fine-tuning the Projection Network.}  The image
adapter model~\cite{ipadapter} takes as input an image,
extracts the corresponding CLIP-embedding and then trains
a projection network $\mathcal{A}$ to learn the mapping into a sequence
of 4 tokens, with dimensions matching the one for text.
In the fine-tuning stage we train the model on the style images
only, for roughly $100$ steps, minimizing the following loss
\begin{equation}
\mathcal{L}_{\mathcal{A}} = \mathop{\mathbb{E}}_{I^s_i\in \mathcal{S}} ||\epsilon - \epsilon^A_\theta(x_t, \tau_\theta(y), t, x^s_i)||^2
\end{equation}
with $\epsilon^A_\theta$ indicating the denoiser consisting
of the original model $\epsilon_\theta$ and the image projection modules.
The model is trained to reconstruct the style images ($I^s_i\in \mathcal{S}$)
, updating only the projection module parameters.

\paragraph{Interpolation of the Style Image Embeddings.}
Fine-tuning helps better capture the style features, but doesn't
address the issue of style and content entanglement.
We observe that the different style images
share the same style while the content varies.
Hence, by averaging the corresponding embeddings we keep the shared property (i.e. the style)
while the differences are toned down (i.e. the content).
During the stylization of any content image $I_c$,
we will use the average features sequence $\phi^s$:
\begin{equation}
\phi^s = \sum_{I^s_i \in\mathcal{S}} \frac{1}{n}\mathcal{A}(I_i^s)
\end{equation}
The middle part of Figure~\ref{fig:method} illustrates
this process.

\subsection{Feature Alignment with Multiple Images}
Although we are able to address content entanglement
thanks to our proposed changes, this is not sufficient to achieve
high quality stylization as the image prompting doesn't capture
all the aspects of the style. 
As a result, we also consider the statistics of the deep
features. Contrary to existing works~\cite{stylealigned,style_inject}
we use multiple images which requires solving a few technical challenges.

\paragraph{Single image style injection and normalization.} 
Let's start with a single style image $I^i_s$,
similarly to existing alignment methods~\cite{style_inject,stylealigned}.
Style injection is achieved by first using DDIM inversion
to obtain the denoising features for a fixed number of steps $T$,
then injecting the features at each time step $t$ during
image stylization.

In the image synthesis pipeline illustrated in Figure~\ref{fig:method}, 
the injection of the style features 
is done in the self-attention layers
during the denoising process.
More specifically, at every self-attention layer and every time step $t$,
we can modify the attention as follows:
\begin{equation}
	\textrm{Attention} (\hat{Q}^t_c, [\hat{K}^t_c~~~K^{i,t}_s], [V_c~~~V^{i,t}_s] ).
\end{equation}

The keys $K^{i,t}_s$ and values $V^{i,t}_s$ are obtained 
during the inversion of the style image $I^i_s$.
Whereas content queries $Q^t_c$ and keys $K^t_c$,
are obtained in the current denoising step $t$,
then normalized using the adaptive normalization~\cite{adain}
\begin{equation}
\hat{Q}^t_c = \textrm{AdaIN}(Q^t_c, Q^{i,t}_s) ~~ \textrm{and} ~~ \hat{K}^t_c = \textrm{AdaIN}(K^t_c, K^{i,t}_s).
\end{equation}
For simplicity, we drop the time step $t$ in the following equations
from our notation as the same operations
are applied independently of $t$. 

\paragraph{Multi-image style injection.} 
A naive extension to multiple images would be
to adjust the attention layer as:
\begin{equation}
	\textrm{Attention} (\hat{Q}_c, [\hat{K}_c~~~K^1_s\ldots K^n_s], [V_c~~~V^1_s \ldots V^n_s] ).
\end{equation}
This is however unfeasible, as the attention values extracted from
the diffusion generation process for a single image over 50 time steps
adds up to almost $7$GB of data, without even considering the
increase in computations.

Given all attention values extracted from all style images,
our solution is to rely on clustering to pick the most unique vectors,
for some lower number of total vectors.
We use KMeans clustering~\cite{kmeans} to cluster values~$\{V^1_s, \ldots, V^n_s\}$.
After clustering, we sample the value vector $v$
closest to each centroid, and select the matching key $k$. The result is a new set of keys $K^\star_s$
and values $V^\star_s$ which represent the style better than a single image while being as compact
\begin{equation}
\textrm{Attention} (\hat{Q}_c, [\hat{K}_c~~~K^\star_s], [V_c~~~V^\star_s] ).
\end{equation}
In practice, we aim to have a number of clusters that matches the target number
of vectors found in a single image (the typical count varies across the UNet),
to compress only the most important and unique style concepts from across
all style images.
This is not a hard limit, and future experiments scaling the number of clusters can be explored.
We perform GPU-accelerated clustering of attention values from each UNet layer,
each timestep, and attention head, separately. This separation enables strong parallelization.

\paragraph{Normalization through an average style image.}
When we scale up to multiple images, the normalization process needs to account
for the attention values from all the style images.
The distribution of attention values 
from multiple images can be multimodal, 
and computing a mean across these different groups of features results
in a value which falls outside the distribution of any individual image,
and produces failed or sub-optimal results.

An effective solution to resolve this issue is to instead 
use an \emph{average} style image generated with 
our multi-image prompt adapter.
Providing only the average style feature embedding $\phi^s$, and no other guidance for
the generation process, produces an \textit{average style image} $\bar{I}^s$ with random content.
The content depicted in $\bar{I}^s$ is unimportant; however
the attention values do fall within the same single distribution,
while covering a wider range of the style (albeit less than in the clustered values).
We can extract the statistics from these attention values,
for use in the normalization and alignment.
\begin{equation}
	\hat{Q}_c = \textrm{AdaIN}(Q_c, \bar{Q}^s) \quad \textrm{and} \quad \hat{K}_c = \textrm{AdaIN}(K_c, \bar{K}^s)
\end{equation}
where $\bar{Q}^s$ and $\bar{K}^s$ the queries and keys
obtained when generating the average style image $\bar{I}^s$, respectively.

In addition to the first two moments, higher orders such as skewness
and kurtosis can be used. 
Finally, we also align the statistics of the latents at each timestep
with the statistics of the latents from $\bar{I}^s$.
This improves style and color quality.

\subsection{High Resolution and Texture Quality}
In addition to the core aspects of the method,
there are a few key points to take into consideration
to achieve best quality results on high resolution output
images.

\begin{figure}[t]
	\centering
	\includegraphics[trim=0 29cm 0 0,clip,width=1\linewidth]{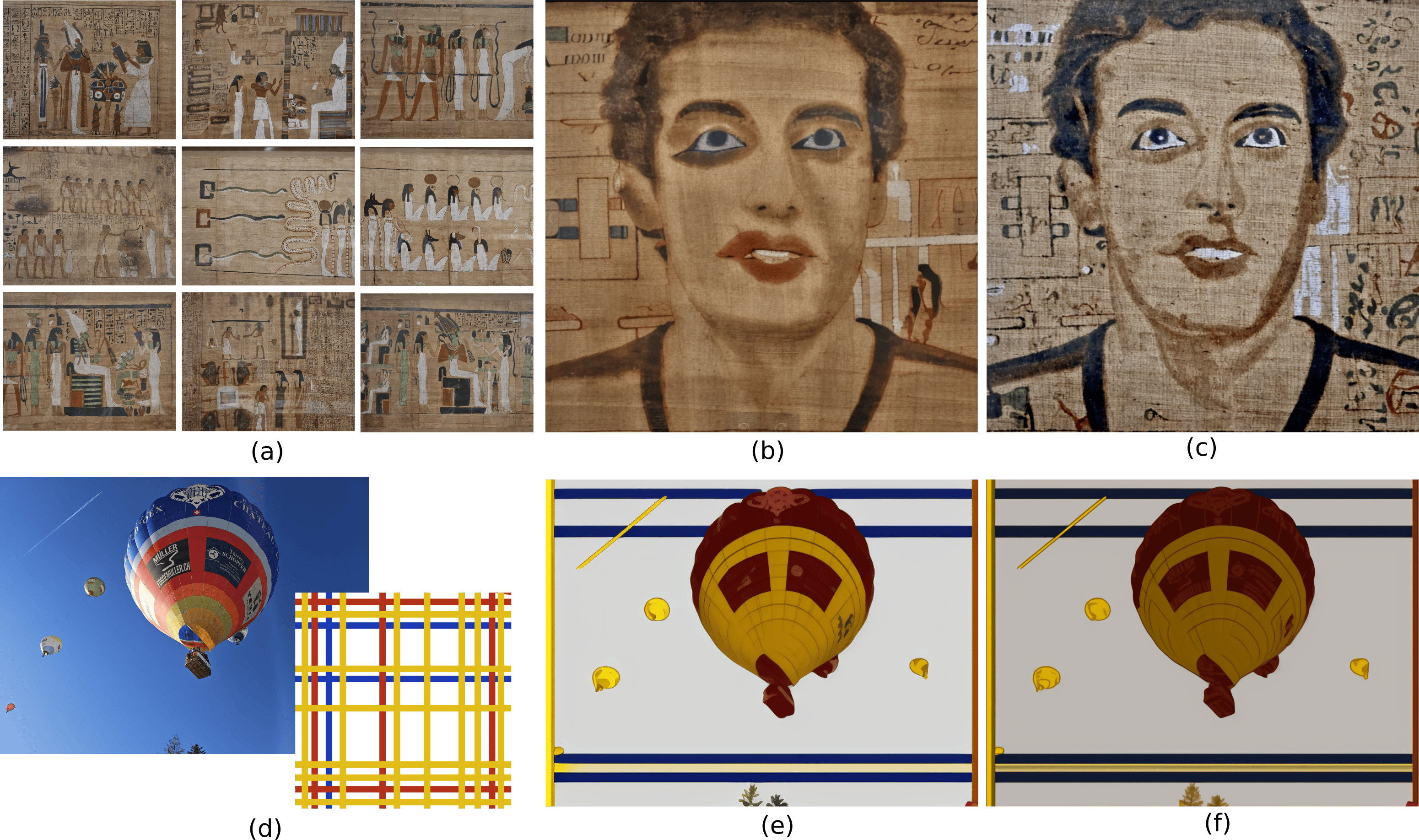}
	\vspace{-0.7cm}
	\caption{Visualization of the effect of our two-stage approach. (a) Starting
		from the set of style images, (b) the first stage (in low resolution)
		produces an initial stylized output however texture details are missing. 
		(c) The second stage helps to add much more detail from the style images.}
	\label{fig:2stage}
	\vspace{-0.4cm}
\end{figure}
\begin{figure}[t]
	\centering
	\includegraphics[trim=0cm 2cm 0cm 5cm,clip,width=1.\linewidth]{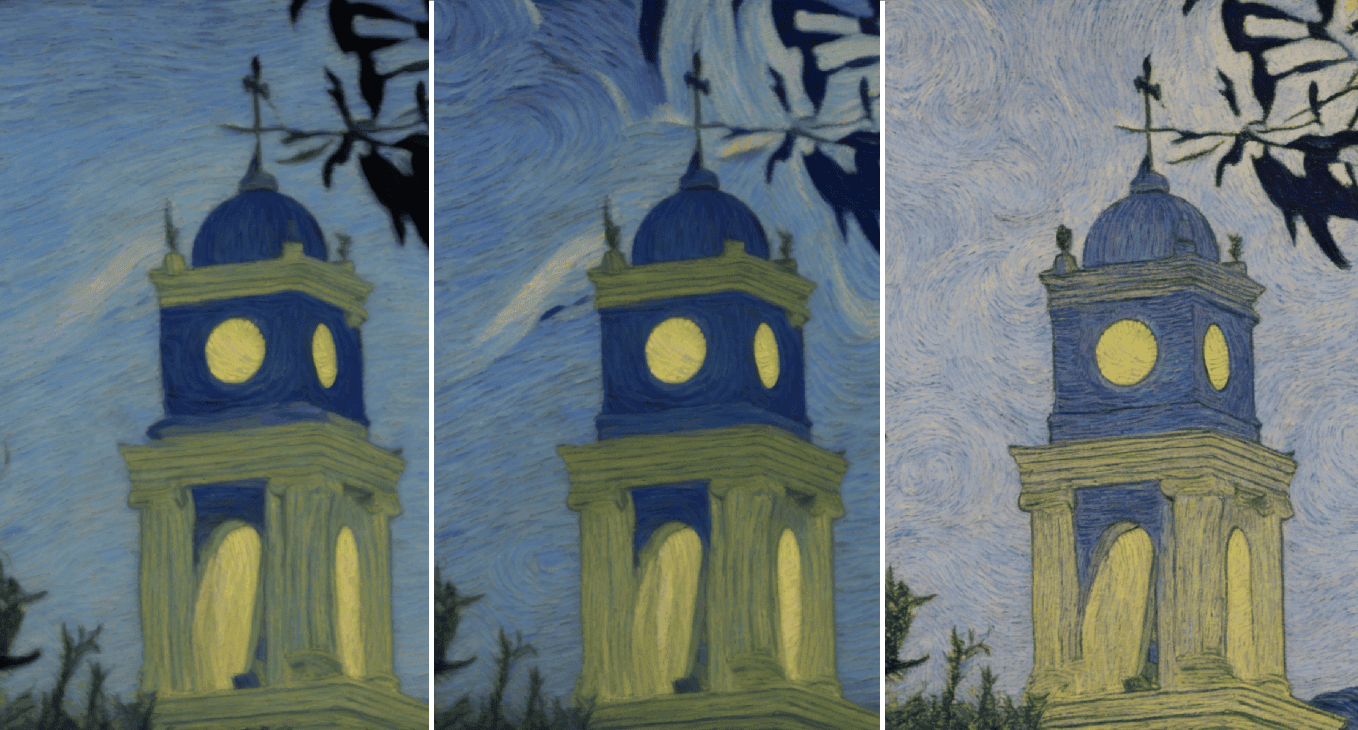}
	\caption{It is possible to control the scale of the textures in the syle transfer output, through scaling
		the style images used in the cross-attention.
		In this example, \textit{Starry Night} painting is used as style source. 
		Changing the scale of the style image crops (from left to right), has a clear effect on 
		the brush strokes texture.
	}
	\label{fig:texture_zoom}
	\vspace{-0.5cm}
\end{figure}

\paragraph{High Resolution Output.}
We use Stable Diffusion as our core model. 
It can be used to generate almost arbitrary resolutions
and aspect ratios. However, scaling the images to higher resolutions,
such as $1024$px and larger, can lead to tiling artifacts
(in part due to the fact that it is trained at a resolution of $512$px).
To avoid this issue, we can rely on the timestep-based 
content disentanglement described by Wu~\emph{et a.}~\cite{diffusiondisentanglement}, 
and focus only the first few timesteps on generating the image structure.
We can use a lower resolution for this, targeting $512$px on the shortest side. 
We can then either spatially resize the latents and resume the rest of the timesteps, or first generate the image in lower resolution then scale up and add details through an image-to-image operation.
As illustrated in Figure~\ref{fig:2stage},
this process avoids tiling artifacts, while allowing the later timesteps to add the lower level stylistic and textural details of a higher resolution image. We use ControlNet depth and LineArt inputs computed separately for the lower and higher resolution content images, during this process, to accurately condition the structure for each appropriate resolution.

\paragraph{Better Texture Quality.}
The CLIP embeddings require the input image to be downsampled,
but stylistic details are lost in such a rescaling operation.
To remedy this issue, we instead opt to compute our final embedding
from a number of patches extracted from the high resolution images.
Thus treating each as a separate style image, in our multi-image pipeline.
In addition to this, the selection of image crops offers a control over the scale of the style
information. Smaller crops can guide the stylization to focus more on the low level textural
details of a style such as brush strokes and lines, whereas larger crops can place more importance
on the structural components of items depicted in the style image.
Fig. \ref{fig:texture_zoom} visualises this effect,
where varying the sizes of the crops can account for such artistic intent.

\begin{figure*}[t]
	\centering
	\includegraphics[width=0.78\linewidth]{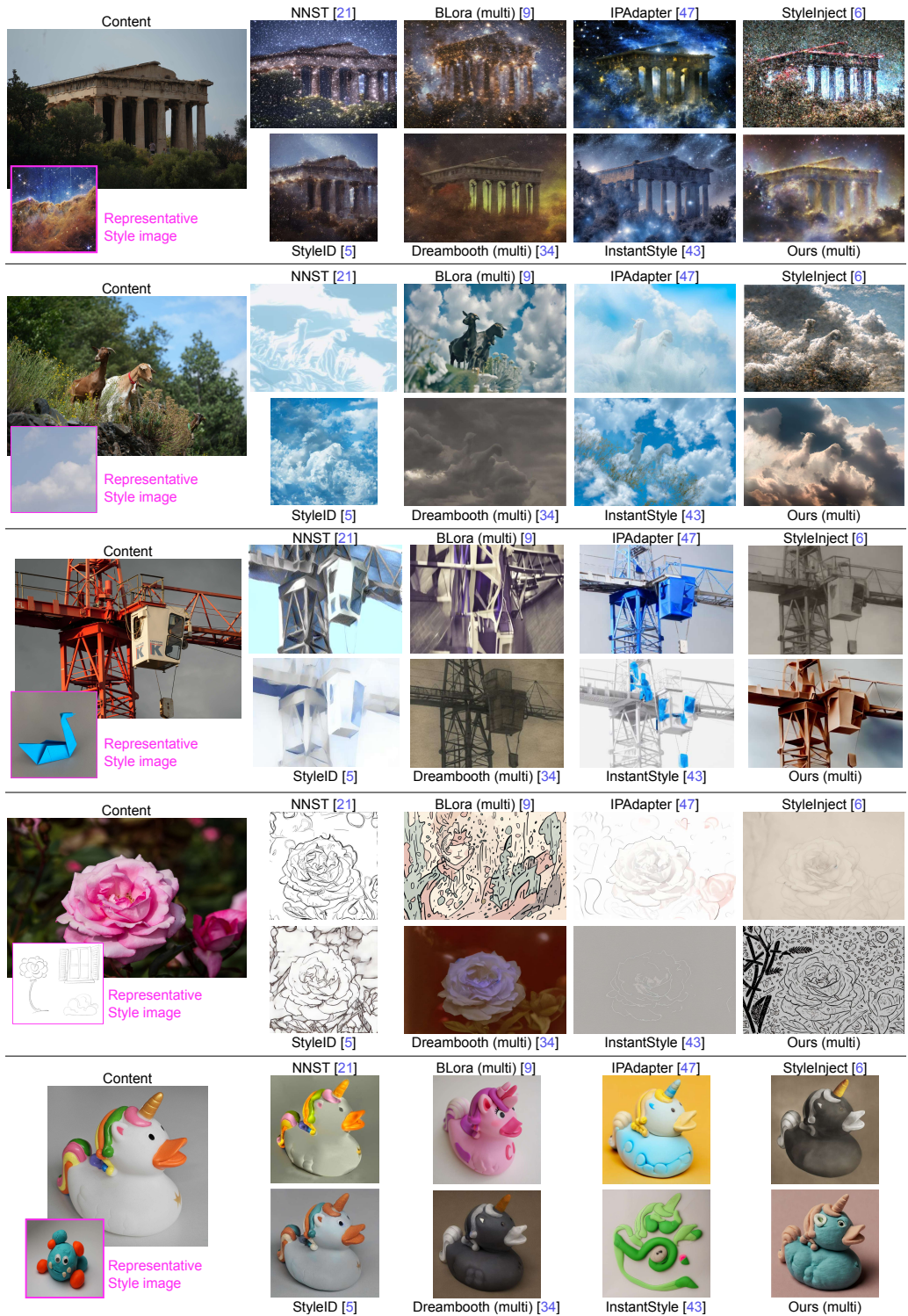}
	\caption{Qualitative results comparison of our method, compared to some of the baselines. 
		Besides ours, we indicate with \emph{(multi)} the methods using multiple images.
		For the methods using a single style image, we highlight with a colored rectangle the style image that was provided.
	}
	\label{fig:results}
	\vspace{-0.29cm}
\end{figure*}

\section{Experiments}
\subsection{Implementation Details.}

We use Stable Diffusion v1.5 for all our experiments, but we show
 in supplementary material that our method generalizes to other backbones. We use the standard ControlNet depth and LineArt.
 Our fine-tuning of the IPAdapter~\cite{ipadapter} takes about 3 to 5 minutes, depending on the number of style images (can be any number). We use the GPU-accelerated Faiss library for clustering, and we implement heavy parallelization, allowing us to run several concurrent instances on the same RTX 4090. This takes under half an hour for a dozen style images and varies depending on their aspect ratios. Once models and data are loaded, stylization on the same machine elapses roughly 16 seconds, also depending on the aspect ratio.

\subsection{Data and Metrics.}
\label{sec:prelim}
We construct a mini dataset for use in testing neural style transfer quality. We form this test set from 50 content images and 200 style images across 15 style groups. Previous style datasets such as BAM \cite{bam}, BBST-4M \cite{neat}, and WikiArt \cite{wikiart}, big or small, are either not licensed openly, not available, or contain style images not grouped into style-consistent groupings. 
We compose this dataset from images that are completely public domain, or from personal images for both content and style sets. 

We use 5 automated metrics for quantitative experiments, computed between each stylized image compared to each of the style images in their respective styles, and averaged. We use SIFID \cite{singan} to measure patch-based style similarity, computed as an FID score between only a pair of images. We use Chamfer distance to measure colour similarity, normalized by the number of pixels to avoid varying image resolution skewing the results. We also use two style embeddings, CSD \cite{csd}, and ALADIN \cite{aladin}, for measuring the similarity of style in a model's embedding space. In all cases a lower value is preferable. Finally, we use similarity in DINOv2 \cite{dinov2} space - this time a higher value is preferable.

\subsection{Comparisons}

We compare against a large range of stylization techniques, some based on diffusion while others are not, 
and some using a single style image, while others can use many. 
Among the techniques not using diffusion we can mention NNST~\cite{nnst}, AdaAttn~\cite{adaattn}
or SANet~\cite{sanet}. 
Among the diffusion methods we compare with IPAdapter~\cite{ipadapter}, 
StyleAligned~\cite{stylealigned} and StyleInject~\cite{style_inject},
and also with other recent works such as InST~\cite{inst} and StyleID~\cite{styleid}.
Some methods are able to leverage multiple style images, 
like Dreambooth~\cite{dreambooth} and BLoRA~\cite{blora}.

For methods using multiple style images (including ours), the entire set is used. 
For methods using a single style image, there is the question of which style to select
for each content. To better represent the performance of these single image style transfer methods,
we consider all possible combination of style and content images. However,
since this would be too computationally involved, we randomly subsample 10\% of these combinations.
The objective is to average out the effect of the particular style image selected,
and better represent the performance of each method.

\begin{table}[t]
	\centering
	\begin{adjustbox}{width=\linewidth}
		\begin{tabular}{llllll}
			Method & SIFID $\downarrow$ & Chamfer $\downarrow$ & CSD $\downarrow$ & A{\footnotesize LADIN} $\downarrow$ & DINO $\uparrow$ \\
			\hline
			IPAdapter~\cite{ipadapter} & 3.427 & 83.632 & 1.091 & 1.109 & 0.654 \\
			StyleAligned~\cite{stylealigned} & 3.988 & 8.716 & 1.326 & 1.362 & 0.597 \\
			NNST~\cite{nnst} & 4.152 & 136.210 & 1.165 & 1.228 & 0.622 \\
			DIFF-NST~\cite{diffnst} & 6.315 & 130.956 & 1.254 & 1.291 & 0.600 \\
			CAST~\cite{cast} & 2.535 & 10.172 & 1.273 & 1.275 & 0.558 \\
			NeAT~\cite{neat} & 2.808 & 8.282 & 1.239 & 1.278 & 0.567 \\
			StyleID~\cite{styleid} & 2.798 & 139.146 & 1.222 & 1.224 & 0.599 \\
			MCCNet~\cite{mccnet} & 2.592 & 6.830 & 1.264 & 1.274 & 0.570 \\
			StyTr2~\cite{stytr2} & 3.136 & 137.742 & 1.202 & 1.213 & 0.616 \\
			PAMA~\cite{pama} & 2.759 & 8.400 & 1.238 & 1.283 & 0.603 \\
			SANet~\cite{sanet} & 3.541 & \textbf{6.289} & 1.227 & 1.252 & 0.601 \\
			AdaAttn~\cite{adaattn} & 3.724 & 10.747 & 1.277 & 1.278 & 0.613 \\
			ContraAST~\cite{contraast} & 2.836 & 6.699 & 1.226 & 1.233 & 0.592 \\
			AdaIN~\cite{adain} & 2.677 & 6.450 & 1.265 & 1.326 & 0.576 \\
			InST~\cite{inst} & 7.032 & 97.497 & 1.179 & 1.227 & 0.614 \\
			S2Wat~\cite{s2wat} & 3.104 & 133.548 & 1.223 & 1.243 & 0.595 \\
			Dreambooth~\cite{dreambooth} & 5.303 & 306.707 & 1.391 & 1.376 & 0.571 \\
			BLoRA~\cite{blora} & 2.870 & 40.799 & 1.120 & 1.118 & 0.658 \\
			BLoRA (multi)~\cite{blora} & 2.512 & 30.257 & 1.119 & 1.097 & 0.666 \\
			InstantStyle~\cite{instantstyle} & 4.479 & 64.331 & 1.150 & 1.107 & 0.668 \\
			\hline
			Ours & \textbf{2.040} & 17.042 & \textbf{1.088} & \textbf{1.054} & \textbf{0.680} \\
		\end{tabular}
	\end{adjustbox}
	\vspace{-0.3cm}
	\caption{Quantitative metrics comparing our method against baselines. Chamfer values are scaled $\times 10^{-3}$ for clarity.}
	\label{tab:baselines}
	\vspace{-0.3cm}
\end{table}

\paragraph{Quantitative Evaluation.}
We present the results of the quantitative evaluation in Table~\ref{tab:baselines}.
We use several metrics (Sec.~\ref{sec:prelim})
to evaluate the performance of the different methods.
From the evaluation it is clear that ours performs best.

It is also interesting to point to the case of BLoRA~\cite{blora},
which can use a single or multiple style images. 
Here the usage of multiple style images improves the performance
which reinforces our message that having access to multiple style images
helps to better capture the style.
In the case of Dreambooth~\cite{dreambooth}, the training
process itself is less stable and no single training setting (learning rate, number of steps, etc.)
is able to achieve good results on all the styles.
For this evaluation we have selected a setting that performs
well on a few style groups and used it for all the rest.
The Chamfer metric measures color matching and we argue it is less important
for the style transfer task. This is confirmed next in the qualitative results
and the user study.

\paragraph{Qualitative Results.}
In Figure~\ref{fig:results} we show a variety of content images
stylized according to different sets of style images. 
Our method is the only approach that performs well
across this wide variety of styles and content.
Using multiple style images helps capture the style,
but this is not sufficient as it can be observed from 
the results of BLoRA (multi)~\cite{blora} or Dreambooth~\cite{dreambooth}.
Of note is the performance of NNST~\cite{nnst}
on styles that are mostly operating on low-level features,
with degradation on examples that need
a better semantic understanding of the content and style.

\begin{figure}[t]
	\centering
	\includegraphics[width=0.9\linewidth]{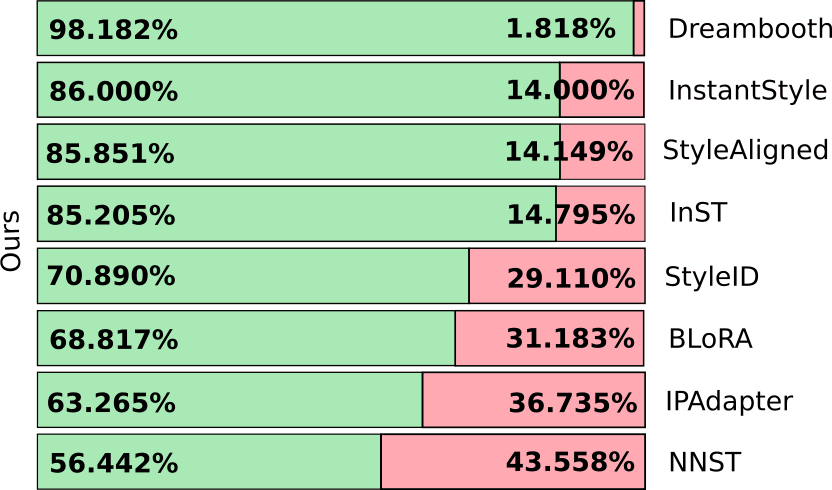}
	\vspace{-0.3cm}
	\caption{Preference scores of our method, compared to each baseline. Besides Ours, Dreambooth and BLoRA use multiple style images.}
	\label{fig:userstudy}
	\vspace{-0.3cm}
\end{figure}

\paragraph{User Study.}

For the user study, we select a subset of the most relevant techniques 
to present them for comparison. 
We select primarily diffusion based techniques, but also NNST~\cite{nnst}, a strong traditional technique.
We present a private team of 23 diverse workers with a labeling task, where a selection of real style images is shown, alongside a shuffled pair of stylization results. One from our own method, and one from a baseline method. We additionally show the real content image being stylized, for context. We ask users to examine the style images, both stylized images, and to select the stylized image which best matches the style for all the real style images shown. In supplementary material, we show a screenshot of an example task shown to a worker. 
Figure~\ref{fig:userstudy} displays the user preference of our method, compared to baselines. 
Our work outperforms all other methods. 
It is interesting to note the good performance of NNST~\cite{nnst} 
in this user study. This illustrates that the metrics (Table~\ref{tab:baselines})
do not cover all aspects of the problem.
Users tend to favor NNST~\cite{nnst}
when the low level features of the style are well preserved,
which works well with many of the styles present in this user study.

\subsection{Ablation Study}

\begin{figure}[t]
	\centering
	\includegraphics[width=1.0\linewidth, trim=50cm 1cm 0cm 5cm,clip]{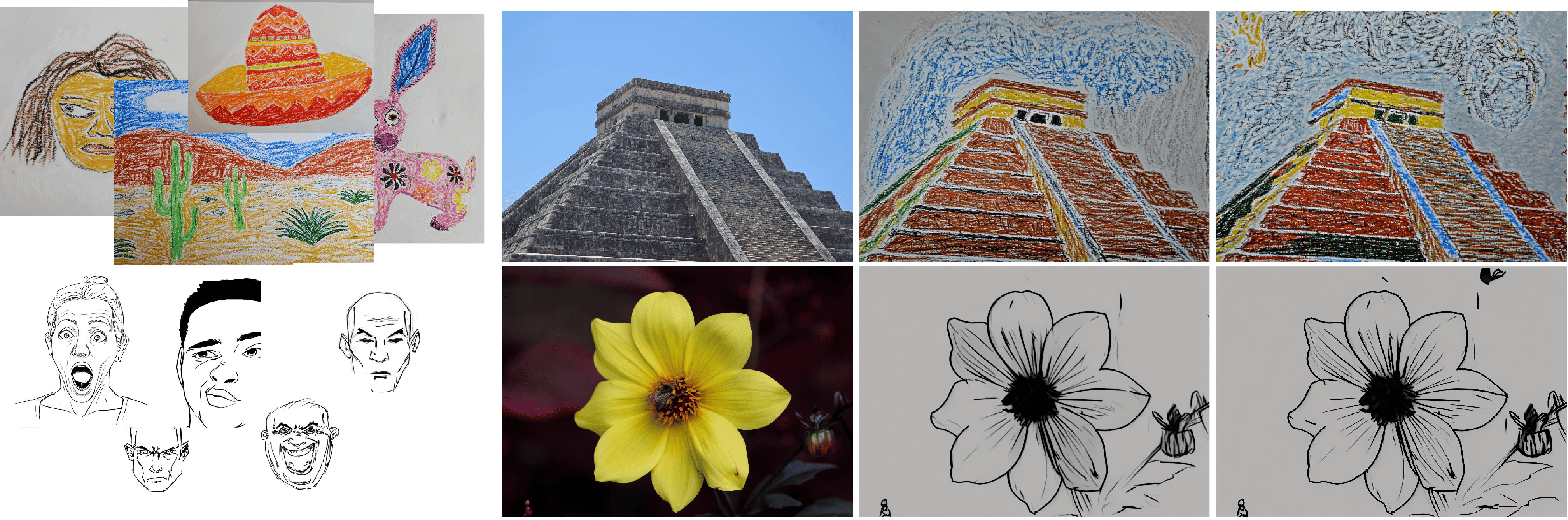}
	\caption{The left-most column shows the content. 
		While being significantly faster ($14s$ \emph{vs} $10$min), 
		the style transfer using attention clustering (middle) 
		has very negligible impact compared to dynamic 
		loading (right-most). }
	\label{fig:non_clustered}
\end{figure}

\paragraph{Clustering.}
As mentioned previously, the clustering step is necessary as the statistical 
alignment process does not scale well, and is limited to around $3$ images
on a single GPU with 24GB of VRAM. 
Still, we would like to evaluate any difference or loss in quality due to using the selected values from
clustering instead of using all the available attention data. 
To make this comparison possible, we apply the stylization using dynamic loading of attention values 
from disk for use in the concatenation step of $\mathcal{K}$ and $\mathcal{V}$ 
self-attention values. 
Of course, this step introduces an extremely unpractical amount of disk reading overhead. 
For a small set of $9$ style images and using an RTX 4090 GPU, this dynamic process 
needs around $9$ minutes and $40$ seconds on average for stylizing one image.
When using clustered values takes around $14$ seconds.
Figure~\ref{fig:non_clustered} shows that the clustering strategy has little effect on the style transfer
results, while being much faster at inference time.

\begin{figure}[t]
	\centering
	\includegraphics[width=0.95\linewidth]{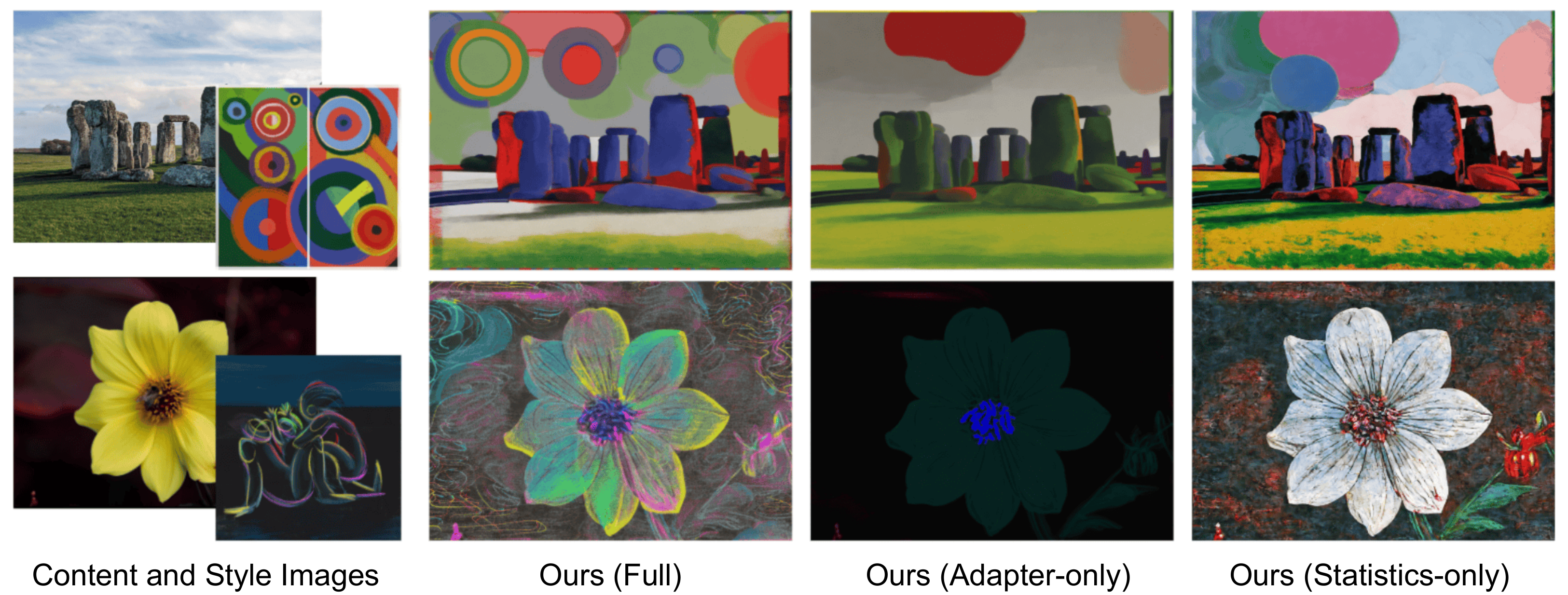}
	\caption{Ablation of cross-attention (adapter-based) and self-attention (statistics-based) components}
	\label{fig:ablate_components}
\end{figure}
\begin{figure}[t]
	\centering
	\includegraphics[width=0.9\linewidth, trim=0cm 0cm 0cm 2cm,clip]{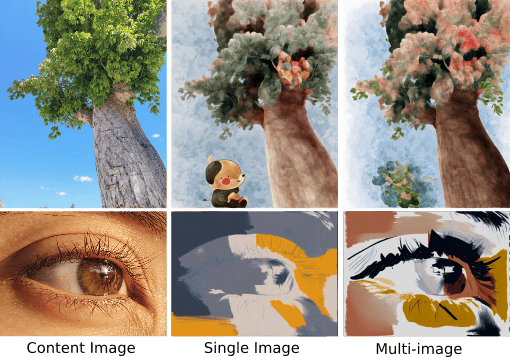}
	\vspace{-0.2cm}
	\caption{Comparison of stylization using either a single source style image, or several. In the single image case, content entanglement can emerge through the erroneous insertion of content from the style images into the results (here, an animal). 
		On the bottom row, we see a far narrower range of colors and brush stroke compared to the range contained in the style group as a whole.}
	\label{fig:single_vs_multi}
\end{figure}

\paragraph{Self-attention vs Cross-attention.}
We describe our method as having two main components. First, cross-attention components, using an image adapter module to inject style features into the diffusion model. Second, a self-attention component using statistics alignment and concatenation on the self-attention values. Both components play a valuable role, as we visualize in Fig \ref{fig:ablate_components}. The statistics component more heavily affects the global appearance of the image, not as heavily dependent on the content, whereas the adapter component more heavily modifies the depicted content. Together, the global style of the image is more thoroughly stylized, while also modifying the content.

\paragraph{Benefits of Using Multiple Style images.}
Our strategy of image adaptation and feature alignment is applicable with single image stylization.
However using multiple style images, is always beneficial (Fig.~\ref{fig:single_vs_multi}). We avoid content entanglement in the results and achieve higher stylization quality.

\paragraph{ControlNet Strength for deformation.}

\begin{figure}[t]
	\centering
	\includegraphics[width=0.85\linewidth]{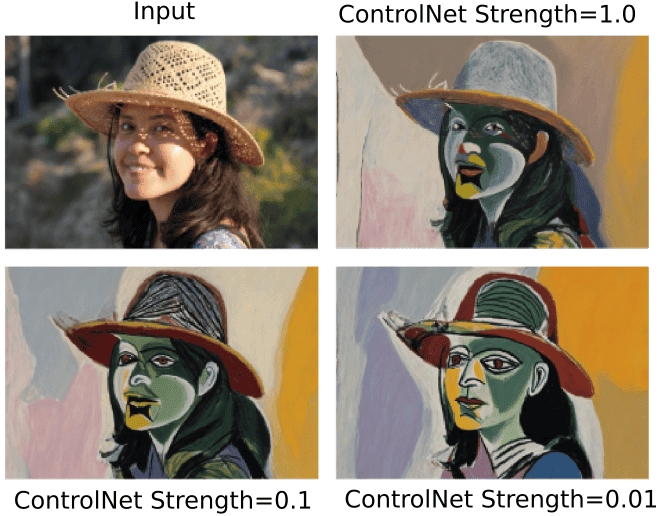}
	\vspace{-0.3cm}
	\caption{The weight of the LineArt ControlNet can be used to control deformation, when used together with depth ControlNet. A lower strength leads to higher deformation, which may be desirable for certain styles.}
	\label{fig:picasso}
	\vspace{-0.4cm}
\end{figure}

ControlNet affects the visual style elements in a stylized image. 
The importance weight of this auxiliary conditioning can expose artistic controls over the stylization process to artists. For example, as shown in Fig \ref{fig:picasso}, a lower strength LineArt ControlNet can result in favorable output when used together with a "deformed" style such as Picasso's cubism.

\section{Discussion}
We propose a model-agnostic diffusion-based style transfer technique that leverages multiple source style images.
We avoid entanglement issues and encode more style variance from a wider range of style examples.
We  show in the quantitative evaluation metrics and user studies that our method is state-of-the-art.
We already show initial results with the larger SDXL~\cite{sdxl} model. Others, such Pixart-$\alpha$(-$\sigma$)~\cite{pixartalpha,pixartsigma},
could be tested.

A key identified limitation of both our technique and the other methods in literature is the lack of control
over more specific or specialized aspects of the stylization, such as line work.
This is an interesting and important future direction of research.

{
    \small
    \bibliographystyle{ieeenat_fullname}
    \bibliography{refs}
}

\end{document}